\documentclass[runningheads]{llncs}
\usepackage[T1]{fontenc}
\usepackage{graphicx,verbatim}
\usepackage{amsmath}
\usepackage{amsfonts}   
\usepackage{amssymb}    
\usepackage{booktabs}
\begin{document}
\title{Hierarchical MoE for Multi-Modal ILD Diagnosis}

\author{
Alec K. Peltekian\inst{1}\orcidID{0000-0001-9082-4000} \and
Gorkem Durak\inst{2}\orcidID{0000-0002-1608-1955} \and
Halil Ertugrul Aktas\inst{2}\orcidID{0009-0009-9491-8122} \and
Carrie Lynn Richardson\inst{3}\orcidID{0000-0002-2541-175X} \and
Mary Carns\inst{4}\orcidID{0000-0002-5063-156X} \and
Kathleen Aren\inst{3}\orcidID{0000-0003-4770-2643} \and
GR Scott Budinger\inst{4,5}\orcidID{0000-0002-3114-5208} \and
Anthony J. Esposito\inst{4,5}\orcidID{0000-0002-8636-0845} \and
Alexander Misharin\inst{4,5}\orcidID{0000-0003-2879-3789} \and
Alok Nidhi Choudhary\inst{1,6}\orcidID{0000-0001-8152-6319} \and
Ankit Agrawal\inst{6}\orcidID{0000-0002-5519-0302} \and
Ulas Bagci\inst{2}\orcidID{0000-0001-7379-6829}
\thanks{Corresponding author.}
}

\authorrunning{A. K. Peltekian et al.}

\institute{
Department of Computer Science, Northwestern University McCormick School of
Engineering and Applied Science, Chicago, IL, USA
\and
Department of Radiology, Machine and Hybrid Intelligence Laboratory,
Northwestern University Feinberg School of Medicine, Chicago, IL, USA
\email{ulas.bagci@northwestern.edu}
\and
Division of Rheumatology, Northwestern University Feinberg School of Medicine,
Chicago, IL, USA
\and
Division of Pulmonary and Critical Care, Northwestern University Feinberg
School of Medicine, Chicago, IL, USA
\and
Simpson Querrey Lung Institute for Translational Science, Northwestern
University Feinberg School of Medicine, Chicago, IL, USA
\and
Department of Electrical and Computer Engineering, Northwestern University
McCormick School of Engineering and Applied Science, Chicago, IL, USA
}
  
\maketitle              
\begin{abstract}
Mixture-of-experts (MoE) models combine specialized predictors under learned routing, offering a principled mechanism for leveraging heterogeneity in medical data. We present a hierarchical multimodal MoE for interstitial lung disease (ILD) classification that integrates a frozen, pre-trained imaging expert with structured electronic health records (EHR) via two-stage gating. A modality-level gate assigns patient-specific weights to imaging and EHR predictions, while a sub-gating module decomposes the EHR branch into clinically defined feature groups with learned, group-specific contributions. This design preserves stable imaging representations while enabling input-dependent clinical weighting and explicit EHR specialization. Under strict patient-level cross-validation, the model achieved the highest mean AUC among the evaluated methods ($0.8750\pm0.0443$), compared with $0.8646$ for imaging-only REN and $0.7685$ for SwinUNETR. The framework extends interpretability across anatomical regions, imaging--EHR utilization, and clinically defined EHR feature groups.
\end{abstract}

%
%
%

\section{Introduction}
Mixture-of-experts (MoE) architectures perform conditional computation by routing inputs to specialized sub-networks~\cite{shazeer2017outrageously,fedus2022switch,mu2025comprehensive}.
Most existing MoE designs assume homogeneous expert capacity and domain-agnostic routing~\cite{riquelme2021scaling,chen2022towards}, which is misaligned with medical imaging, where anatomical structure and region-specific pathology impose strong constraints on expert specialization and require interpretable, domain-aware modeling~\cite{litjens2017survey,dalca2018anatomical}.

Pulmonary imaging illustrates this mismatch. Interstitial lung disease (ILD) exhibits heterogeneous, regionally distributed involvement; clinical interpretation of high-resolution CT is inherently regional, with patterns such as basal and subpleural predominance~\cite{raghu2018diagnosis,desai2004ct}.
Despite this, most deep-learning approaches process the lung as a single global volume, diluting regional signals and limiting interpretability~\cite{humphries2024deep}. Anatomically structured MoE models have improved ILD classification by promoting region-aware specialization~\cite{peltekian2025ren}, yet remain imaging-only and do not incorporate the structured clinical information that is central to ILD diagnosis in practice~\cite{chatterjee2023multidisciplinary}.

Integrating imaging and structured EHR can be formally viewed as a heterogeneous conditional risk minimization problem in which the optimal predictive function depends on patient-specific modality relevance. Let $X_I$ denote imaging features and $X_E$ structured clinical variables. The Bayes-optimal decision rule for ILD diagnosis is generally not separable across modalities and may vary across subpopulations. Fixed fusion strategies implicitly assume uniform modality utility, resulting in suboptimal bias when modality informativeness is patient-dependent. We therefore formulate multimodal ILD classification as a structured conditional computation problem, where routing mechanisms approximate a patient-specific decomposition of the predictive function:
\begin{equation}
f(X_I, X_E) = \sum_{m} g_m(X_I, X_E)\, f_m(X_I, X_E),
\end{equation}
where gating functions $g_m$ dynamically allocate representational capacity across modality- and domain-specialized experts. Under this perspective, effective multimodal modeling requires not only feature fusion but principled routing that reflects anatomical and clinical structure.

The two modalities encode complementary but heterogeneous information whose relative diagnostic value varies across patients, comorbidities, and disease stages~\cite{huang2020fusion,mohsen2022artificial}.
Prior multimodal approaches have explored feature concatenation, attention-based fusion, and cross-modal transformers~\cite{li2024review,kamrul2025multimodal}, yet typically rely on dense interaction or fixed fusion strategies without explicit, structured routing across modalities or clinically defined feature groups. In some cases, radiologic findings alone suffice to exclude disease; in others, subtle impairments in pulmonary function or clinical phenotype provide the decisive signal. Effective multimodal models must therefore adaptively weight information across modalities and clinical domains while maintaining the interpretability required for clinical deployment.

\begin{figure*}[t]
    \centering
    \includegraphics[width=\textwidth]{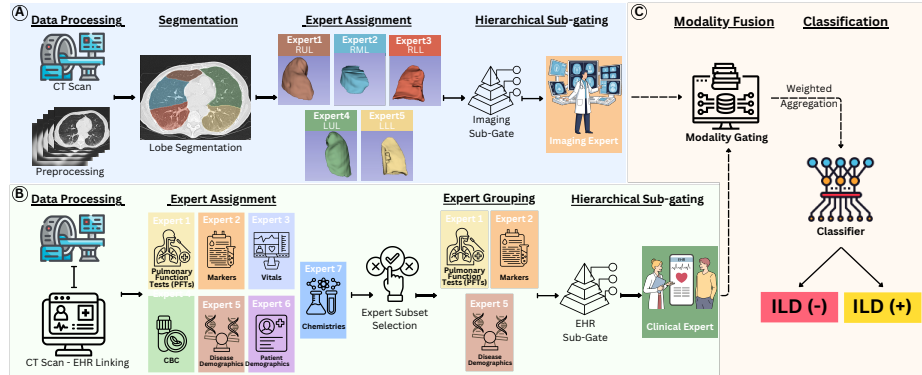}
    \caption{
    Overview of the proposed hierarchical multimodal MoE framework for ILD diagnosis.
    \textbf{(A)} CT pipeline: volumes are preprocessed and segmented into five lung lobes; five lobe-specific imaging experts operate on lobe-masked volumes; fixed radiomics-derived lobe-importance weights provide a fold-specific prior for aggregating expert representations.
    \textbf{(B)} EHR pipeline: structured variables are partitioned into clinically defined feature groups and processed by group-specific sub-experts; an EHR sub-gate aggregates sub-expert outputs into a single clinical representation. We additionally evaluate a \emph{selective} EHR configuration via expert-subset selection (restricting to a targeted subset of feature groups) prior to EHR sub-gating.
    \textbf{(C)} Multimodal fusion: a modality gate adaptively weights imaging and clinical representations, followed by a classifier for ILD prediction.
    }
    \label{fig:architecture}
\end{figure*}

We introduce a hierarchical multimodal MoE architecture for ILD classification with two-level gating (Fig.~\ref{fig:architecture}). A \emph{modality-level gate} assigns patient-specific weights to an imaging expert and an EHR branch, enabling adaptive reliance on each modality. Within the EHR branch, a \emph{sub-gating module} decomposes structured clinical variables into clinically defined feature groups and learns group-specific contributions, providing structured, input-dependent fusion while preserving interpretability. We evaluate the model under patient-level cross-validation and achieve the highest mean AUC among the evaluated methods, together with case-level summaries of modality and clinical-feature usage derived from learned gate activations.

\section{Methods}
\textbf{Dataset and Preprocessing.} We used a retrospective cohort from the Northwestern Scleroderma Registry: 597 patients with 1{,}898 longitudinal chest CT scans (2001--2023). The cohort comprised 489 (81.9\%) female patients (mean age 63.7$\pm$12.7\,yr). Disease subtypes included limited cutaneous SSc (47.6\%), diffuse cutaneous SSc (41.0\%), SSc sine scleroderma (4.7\%), and other (6.7\%). ILD was confirmed in 365 patients (61.1\%), forming the positive class.

CT volumes were resampled to isotropic 1\,mm spacing, segmented into five anatomical lung lobes (LUL, LLL, RUL, RML, RLL) using LungMask R231~\cite{hofmanninger2020automatic}, intensity-windowed to $[-175,250]$\,HU, and resized to $96{\times}96{\times}96$ voxels. The study was approved by the Northwestern IRB with informed consent from all participants.

\textbf{Radiomics-Guided Lobe Importance.} Following anatomically informed MoE work~\cite{peltekian2025ren}, we estimate lobe-level diagnostic importance via radiomics. For each cross-validation fold, 107 PyRadiomics features~\cite{van2017computational} are extracted per lobe, and lobe-specific XGBoost classifiers~\cite{chen2016xgboost} are trained on the training split. Validation AUCs are converted into fixed attention weights:
\begin{equation}
  w_k = \begin{cases} 0.1 + (\mathrm{AUC}_k - 0.5)\cdot 3.8, & \mathrm{AUC}_k \geq 0.5 \\ 0.1, & \text{otherwise} \end{cases}
  \label{eq:lobe_weights}
\end{equation}
This mapping assigns a minimum nonzero contribution to lobes with validation AUC below chance and increases the weight linearly for more predictive lobes. The weights are computed once per fold and held fixed during multimodal training. Alternative mappings were not systematically evaluated.

\textbf{Lobe-Aware Imaging Expert.} The imaging branch consists of five lobe-specific experts (LUL, LLL, RUL, RML, RLL), each implemented using a 3D SwinUNETR backbone~\cite{hatamizadeh2021swin}. Lobe masks restrict feature extraction to anatomically meaningful regions. Each lobe expert produces a representation $\boldsymbol{\phi}_{\mathrm{img},k}$, and these are aggregated using the fixed radiomics-derived lobe importance weights from Eq.~\eqref{eq:lobe_weights}.

This fold-specific weighting can be interpreted as a data-dependent prior over expert relevance, introducing a structured inductive bias into the routing process. By decoupling anatomical importance estimation from gradient-based optimization, we constrain the imaging MoE to regionally consistent solutions while preserving specialization. Formally, this implements a constrained optimization:
\begin{equation}
\min_{\theta} \mathcal{L}\big(f_{\theta}\big)
\quad
\text{s.t.}
\quad
w_k = h(\mathrm{AUC}_k),
\end{equation}
where $h(\cdot)$ is the lower-bounded affine mapping defined in 
Eq.~\eqref{eq:lobe_weights}. The resulting weights are normalized before expert 
aggregation:
\begin{equation}
\tilde{w}_k = \frac{w_k}{\sum_{j=1}^{5} w_j}, \qquad
\boldsymbol{\phi}_{\mathrm{img}} = \sum_{k=1}^{5} \tilde{w}_k \boldsymbol{\phi}_{\mathrm{img},k}.
\end{equation}

Global average pooling and a projection head yield a 48-dimensional imaging representation $\boldsymbol{\phi}_{\mathrm{img}} \in \mathbb{R}^{48}$. The imaging expert is trained independently on CT data and \emph{frozen} during multimodal training, preserving lobe-specific specialization while the EHR branch and gating networks are optimized.

\textbf{EHR Processing and Feature Grouping.}
Structured EHR variables were linked separately to each CT examination. For each variable, the most recent measurement recorded on or before the CT date was selected. When no prior measurement was available, the nearest subsequent measurement within a 30-day grace window was permitted. Values unavailable within this interval were zero-filled before standardization.

EHR features were standardized independently within each cross-validation fold using feature-wise means and standard deviations computed exclusively from the training partition. The same transformation was applied unchanged to the validation and test partitions. Explicit missingness indicators were not used.

To enable hierarchical specialization, EHR variables were partitioned into clinically coherent groups.
We evaluated two configurations:
\textbf{(i) Full hierarchy (7 groups; 69 variables total).}
EHR variables were partitioned into seven clinically coherent groups:
Complete Blood Count (21 variables), Pulmonary Function Tests (3 variables),
Serum Chemistries (14 variables), Laboratory and Functional Markers (4 variables),
Vital Signs (7 variables),
Static Patient Demographics (14 variables), and
Disease-Specific Demographics (6 variables).

\textbf{(ii) Selective hierarchy (3 groups; 12 variables total).}
The selective configuration restricts the EHR branch to three clinically proximal groups:
Pulmonary Function Tests (FVC, FEV1, DLCO),
Disease-Specific Demographics (systemic sclerosis subtype indicators [lcSSc, dcSSc, SSS] and autoantibody status including Scl-70, anticentromere antibody, and RNA polymerase III),
and Key Biomarkers (peak creatinine, modified Rodnan skin score, and B-type natriuretic peptide).
This selective grouping concentrates routing capacity on dynamic and mechanistically relevant predictors while excluding broader static demographic variables.

\textbf{Hierarchical EHR Expert.} Grouping EHR variables into clinically coherent subsets imposes a structured sparsity prior over multimodal interactions. Without grouping, dense fusion induces high-variance cross-modal entanglement, especially when weakly informative or static variables are present. Let $X_E=\{X_1,\ldots,X_G\}$ denote grouped features. The hierarchical MoE approximates:
\begin{equation}
f_E(X_E)= \sum_{g=1}^{G} \pi_g(X_E)\, f_g(X_g),
\end{equation}
where $f_g$ are group-specific sub-experts and $\pi_g$ are input-dependent routing weights. This decomposition reduces the effective interaction dimensionality from $\mathcal{O}(|X_E|^2)$ to $\mathcal{O}(G)$, controlling variance while preserving conditional flexibility.

Let $\mathbf{x}_g$ denote the features belonging to group $g$.
Each group is processed by a dedicated sub-expert with residual projection:
\begin{equation}
  \boldsymbol{\phi}_g = \mathrm{MLP}_g(\mathbf{x}_g) + \mathrm{Proj}_g(\mathbf{x}_g),\quad \boldsymbol{\phi}_g \in \mathbb{R}^{24}.
  \label{eq:sub_expert}
\end{equation}
Sub-expert outputs are aggregated via a learned gating network:
\begin{equation}
  \mathbf{g}_{\mathrm{sub}} = \mathrm{Softmax}\!\bigl(\mathrm{Gate}_{\mathrm{sub}}([\boldsymbol{\phi}_1;\ldots;\boldsymbol{\phi}_G])\bigr).
  \label{eq:sub_gate}
\end{equation}
The final EHR representation combines gated and direct information through a dual-fusion strategy:
\begin{equation}
  \boldsymbol{\phi}_{\mathrm{ehr}} = \mathrm{Fusion}([\boldsymbol{\phi}_1;\ldots;\boldsymbol{\phi}_G]) + \mathrm{Proj}\!\Bigl(\textstyle\sum_{g=1}^{G} g_{\mathrm{sub},g}\,\boldsymbol{\phi}_g\Bigr),
  \label{eq:ehr_fusion}
\end{equation}
yielding $\boldsymbol{\phi}_{\mathrm{ehr}} \in \mathbb{R}^{48}$.

\textbf{Modality-Level Gating and Classification.}
A modality gate adaptively balances imaging and EHR contributions:
\begin{equation}
  \mathbf{g}_{\mathrm{mod}} = \mathrm{Softmax}\!\bigl(\mathrm{Gate}_{\mathrm{mod}}([\boldsymbol{\phi}_{\mathrm{img}};\boldsymbol{\phi}_{\mathrm{ehr}}])\bigr) \in \mathbb{R}^{2}.
  \label{eq:mod_gate}
\end{equation}
The fused representation $\boldsymbol{\phi}_{\mathrm{final}} = g_{\mathrm{mod},1}\,\boldsymbol{\phi}_{\mathrm{img}} + g_{\mathrm{mod},2}\,\boldsymbol{\phi}_{\mathrm{ehr}}$ is passed to a lightweight MLP classifier for binary ILD prediction.

\textbf{Training Protocol.}
Training proceeds in three stages: (1) radiomics-based lobe importance estimation, (2) imaging expert training with lobe-aware attention, and (3) multimodal training with the imaging expert frozen while EHR sub-experts, gating networks, and the classifier are optimized jointly. All models use AdamW (lr\,$=$\,$4{\times}10^{-4}$, weight decay $10^{-5}$)\cite{loshchilov2017decoupled}, batch size 4, early stopping (patience 10), and inverse-frequency class weighting. Evaluation uses strict patient-level 5-fold cross-validation, with all longitudinal CT scans from each patient retained in the same fold to prevent patient-level leakage. AUC is reported as mean$\pm$SD across folds. Statistical comparisons used two-sided paired $t$-tests on fold-level AUCs obtained from identical patient-level partitions; the selective hierarchical MoE was additionally compared directly with imaging-only REN. Data augmentation includes random rotation ($\pm$15$^{\circ}$), scaling (0.9--1.1), and Gaussian noise. All models were implemented in PyTorch/MONAI on NVIDIA A100 GPUs.

\section{Results}
Table~\ref{tab:results} summarizes performance across all methods. The 
selective hierarchical MoE achieved the highest mean AUC 
($0.8750\pm0.0443$), compared with $0.8646\pm0.0467$ for imaging-only REN, corresponding to an absolute AUC increase of $0.0104$. However, the paired fold-level difference was not statistically significant 
($t(4)=0.265$, $p=0.804$), so the numerical improvement requires confirmation in a larger evaluation. The selective model significantly outperformed SwinUNETR ($p<0.001$). Notably, the highest mean AUC was achieved with the imaging expert frozen while only the EHR branch, gating networks, and final classifier were optimized.

\begin{table*}[!t]
\centering
\caption{Performance comparison using 5-fold patient-level cross-validation. Values are reported as mean$\pm$SD across folds. Table $p$-values are from two-sided paired $t$-tests against SwinUNETR. The direct comparison between the selective hierarchical MoE and imaging-only REN is reported in the text.}
\label{tab:results}
\resizebox{\textwidth}{!}{%
\begin{tabular}{lccc}
\toprule
\textbf{Method} & \textbf{AUC $\pm$ SD [95\% CI]} & \textbf{Change vs. SwinUNETR} & \textbf{p-value} \\
\midrule

\multicolumn{4}{l}{\textit{Imaging-Only Baselines}} \\
SwinUNETR (Baseline) 
& $0.7685 \pm 0.0759$ [0.674, 0.863] 
& -- 
& -- \\

CNN (Baseline) 
& $0.7584 \pm 0.0919$ [0.667, 0.850] 
& $-1.3\%$ 
& -- \\

Mamba (Baseline) 
& $0.6775 \pm 0.0454$ [0.632, 0.723] 
& $-11.8\%$ 
& -- \\

ViT (Baseline) 
& $0.6535 \pm 0.0356$ [0.618, 0.689] 
& $-15.0\%$ 
& -- \\

\midrule
\multicolumn{4}{l}{\textit{MoE Anatomically-Guided Imaging-Only (REN)}} \\
Radiomics-Guided REN 
& $0.8646 \pm 0.0467$ [0.806, 0.923] 
& $+12.5\%$ 
& 0.031 \\

\midrule
\multicolumn{4}{l}{\textit{Multi-Modal Approaches}} \\

Hierarchical MoE (Full EHR) 
& $0.8496 \pm 0.0449$ [0.794, 0.905] 
& $+10.6\%$ 
& 0.004 \\

\textbf{Hierarchical MoE (Selective EHR)} 
& $\mathbf{0.8750 \pm 0.0443}$ [0.820, 0.930]
& $\mathbf{+13.9\%}$ 
& $\mathbf{<0.001}$ \\

\bottomrule
\end{tabular}%
}
\end{table*}

\textbf{Effect of Hierarchical EHR Grouping.}
The selective hierarchy achieved a higher mean AUC than the full hierarchy ($0.8750$ vs.\ $0.8496$). A plausible explanation is that the full configuration distributes routing capacity across 69 variables and seven groups, including static or potentially redundant variables. The selective hierarchy instead concentrates routing on pulmonary function, disease phenotype and serology, and key biomarkers that are more proximal to ILD status. This interpretation is consistent with the concentration of selective sub-gate weights on the PFT expert, although it does not establish that the excluded variables are intrinsically uninformative.

\textbf{Robustness, longitudinal stability, and fusion baselines.}

\underline{Temporal linkage sensitivity:}
Varying the forward grace window used for EHR--CT matching when no prior measurement existed (0, 7, 14, 30, 60, and 90 days) produced negligible changes in performance (AUC $0.8746 \pm 0.0443$, $0.8746 \pm 0.0443$, $0.8749 \pm 0.0442$, $0.8750 \pm 0.0443$, $0.8750 \pm 0.0443$, and $0.8750 \pm 0.0441$, respectively), indicating robustness to reasonable temporal linkage policies.

\underline{Longitudinal prediction variance:}
Among patients with more than one CT scan ($n=192$), the mean within-patient standard deviation of predicted ILD probability was $0.0812 \pm 0.0964$, with a mean within-patient probability range of $0.2004 \pm 0.2406$.
Variability was slightly lower in ILD-positive patients ($n=138$; standard deviation $0.0756 \pm 0.0922$, range $0.1901 \pm 0.2344$) than in ILD-negative patients ($n=54$; standard deviation $0.0953 \pm 0.1060$, range $0.2265 \pm 0.2562$), consistent with higher uncertainty in borderline-negative cases.

\underline{Multimodal integration and late-fusion baselines:}
Under identical patient-level folds, the selective hierarchical MoE achieved $0.8750\pm0.0443$ AUC, compared with $0.8646\pm0.0467$ for imaging-only REN. Among simpler integration strategies, concatenation followed by logistic regression achieved $0.8595\pm0.0438$, concatenation with an MLP achieved $0.8337\pm0.0551$, and EHR-only logistic regression achieved $0.8280\pm0.0517$. These results show that the complete hierarchical model achieved the highest mean AUC among the evaluated fusion strategies, although they do not independently isolate the contributions of the modality gate and EHR sub-gate.

\begin{figure*}[t]
    \centering
    \includegraphics[width=0.9\textwidth]{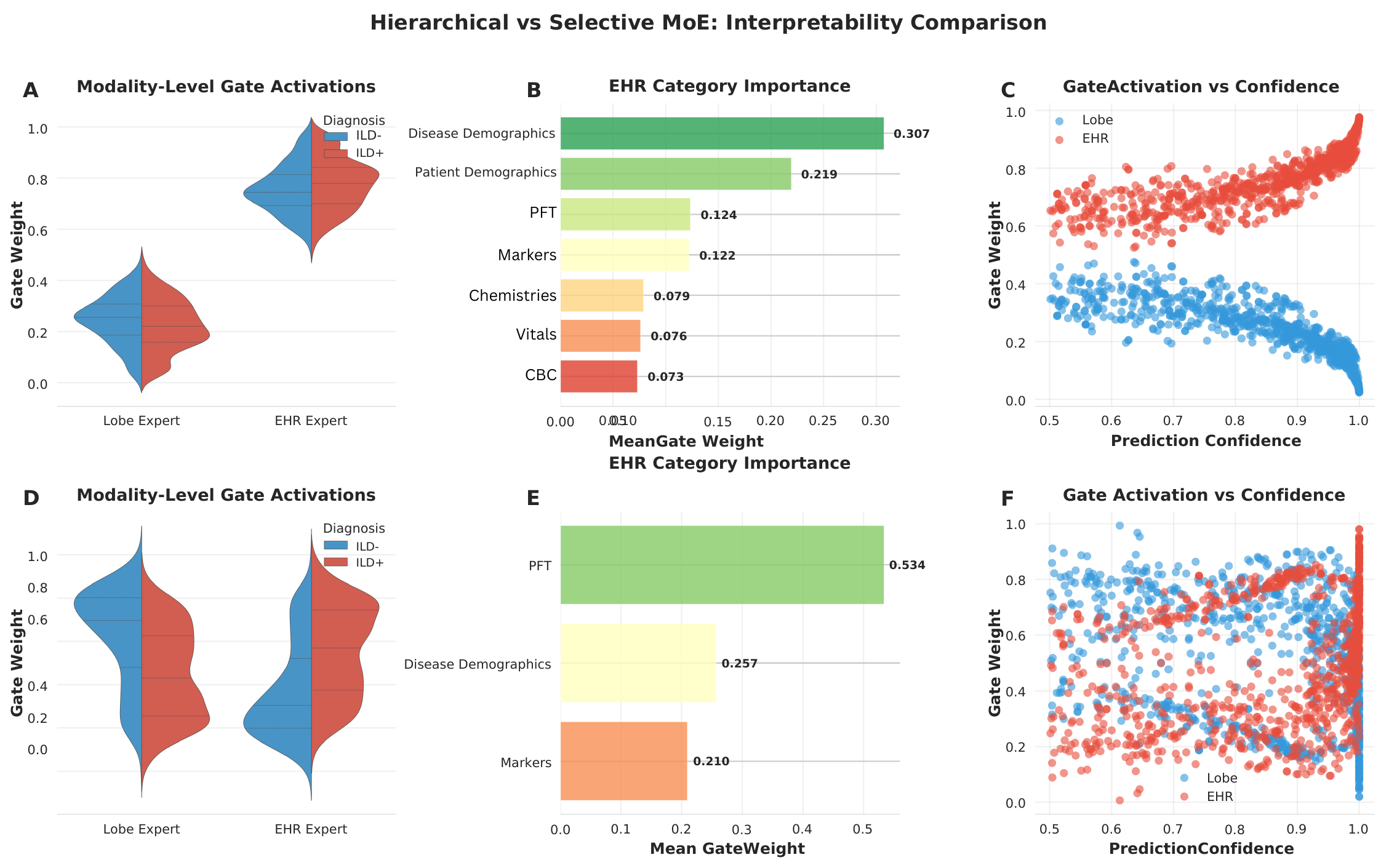}
    \caption{
    Interpretability analysis of hierarchical MoE gating behavior.
    (A) Modality-level gate activations for the full hierarchical EHR configuration, stratified by ILD status.
    (B) Mean EHR sub-expert gate weights across clinical feature groups for the full hierarchy.
    (C) Relationship between modality-level gate activations and prediction confidence for the full hierarchy.
    (D-F) Corresponding analyses for the selective hierarchical EHR configuration.
    }
    \label{fig:gating}
\end{figure*}

\textbf{Interpretability of Hierarchical Gating.}
We analyzed learned gate activations at both the modality and EHR sub-expert levels (Fig.~\ref{fig:gating}). Gate activations represent model routing behavior rather than causal feature importance. A high gate weight indicates greater utilization of an expert but does not establish its independent clinical importance.

\underline{Modality-level routing:}
Imaging experts receive higher weights for ILD-negative predictions, consistent with clear parenchyma being sufficient to exclude disease. EHR experts are weighted more heavily for ILD-positive predictions, particularly at higher confidence, mirroring clinical workflows in which confirmation of ILD relies on pulmonary function impairment and disease phenotype.

\underline{EHR sub-expert routing:}
In the full hierarchy, routing is distributed broadly across clinical categories, with increasing EHR dominance at high confidence---a pattern associated with overconfident positive predictions.
In the selective hierarchy, routing concentrates on PFT sub-experts, followed by disease demographics and biomarkers, while modality-level gating maintains more balanced imaging--EHR usage across confidence levels.
Together, these patterns provide a model-level interpretation of the observed performance difference: restricting EHR integration to clinically proximal groups reduces excessive EHR dominance, preserves complementary imaging evidence, and improves calibration.

\underline{Calibration and Error Analysis:}
Across five folds, the selective hierarchical model achieved an ECE of $0.131\pm0.038$ and a Brier score of $0.135\pm0.037$, indicating reasonable calibration. Of 960 test samples, 167 (17.4\%) were misclassified, with 71 errors (42.5\%) at high confidence ($\geq$0.8). Misclassified cases exhibited lower EHR reliance (mean gate weight 0.383 vs. 0.527 for correct predictions), while false positives were more strongly imaging-driven. These patterns suggest high-confidence errors are associated with imaging-dominant routing in clinically ambiguous cases, motivating confidence-aware routing in future work.

\underline{Modality Contribution Analysis:}
The selective hierarchical MoE achieves balanced average modality weights (imaging 0.498$\pm$0.122, EHR 0.502$\pm$0.122), contrasting with fixed or late-fusion strategies that assume static contributions. Learned gates adaptively emphasize imaging or EHR per sample, providing a soft, patient-specific alternative to non-hierarchical fusion.

\section{Discussion and Concluding Remarks}
We introduced a hierarchical multimodal MoE framework that extends anatomically informed REN with interpretable routing across imaging, EHR, and clinically defined feature groups. The selective configuration achieved the highest mean AUC while providing case-specific summaries of anatomical, modality, and EHR-group utilization. Although the numerical improvement over imaging-only REN was not statistically significant, the results motivate further evaluation of clinically structured multimodal routing in larger and external cohorts.

\textbf{Limitations and future work.}
Our cohort originates from a single academic health system and focuses on 
scleroderma-associated ILD, which may limit generalizability to other 
etiologies. However, patients were imaged across multiple affiliated hospitals with diverse scanners over two decades, reducing the likelihood that performance reflects a narrow acquisition pipeline. EHR feature grouping relies on clinically motivated rather than learned partitions, and data-driven groupings may capture additional interactions. The framework also depends on automated lung and lobe segmentation, introducing potential error propagation. Missing EHR values were represented without explicit missingness indicators, which may limit the model's ability to distinguish unavailable measurements from observed values after preprocessing. Furthermore, the current comparisons 
do not independently isolate the contributions of the modality gate and EHR sub-gate. Future work should evaluate explicit missingness modeling, 
component-wise gating ablations, patient-level endpoints, multi-institutional external validation, longitudinal progression and subtype-aware prediction, and model compression for deployment. Although developed for systemic sclerosis-associated ILD, the framework may also be applicable to other multimodal diagnostic tasks involving structured imaging and clinical data.

\bibliographystyle{splncs04}
\bibliography{references}
\end{document}